%% file: bare_jrnl_new_sample4.tex
\documentclass[lettersize,journal]{IEEEtran}
\usepackage{amsmath,amsfonts}
\usepackage{algorithmic}
\usepackage{algorithm}
\usepackage{array}
\usepackage[caption=false,font=normalsize,labelfont=sf,textfont=sf]{subfig}
\usepackage{textcomp}
\usepackage{stfloats}
\usepackage{url}
\usepackage{verbatim}
\usepackage{graphicx}
\usepackage{color}
\usepackage{float} 
\usepackage{booktabs}
\usepackage{cite}
\usepackage{subfig}
\usepackage{caption}
\usepackage{multirow}
\usepackage{bbding}
\usepackage{tabularray}
\usepackage{colortbl}
\begin{document}

\title{Improving Multi-Person Pose Tracking \\with A Confidence Network}

\author{Zehua~Fu\dag,
        Wenhang~Zuo\dag,
        Zhenghui~Hu,
        Qingjie~Liu*,~\IEEEmembership{Member,~IEEE},
        Yunhong~Wang,~\IEEEmembership{Fellow,~IEEE}
\thanks{\dag \ These authors contribute equally to this work and should be considered co-first authors. * Corresponding author. \\ \indent
Wenhang Zuo, Qingjie Liu and Yunhong Yang are with the State Key Laboratory of Virtual Reality Technology and Systems, School of Computer Science and Engineering, Beihang University, Beijing 100191, China. E-mail:~zuowenhang@gmail.com, \{qingjie.liu, yhwang\}@buaa.edu.cn. \\ \indent
Zehua Fu and Zhenghui Hu are with Hangzhou Innovation Institute, Behang University,
Hangzhou 310051, China. Email:~\{zehua\_fu, zhenghuihu2021\}@buaa.edu.cn.

}
}

\markboth{Journal of \LaTeX\ Class Files,~Vol.~14, No.~8, August~2021}%
{Shell \MakeLowercase{\textit{et al.}}: A Sample Article Using IEEEtran.cls for IEEE Journals}


\maketitle

\begin{abstract}
\input{sections/abstract.tex}    
\end{abstract}

\section{Introduction}
\input{sections/introduction}
\section{Related Work}
\input{sections/related_work}
\section{Methodology}
\input{sections/methodology}
\section{Experiments}
\input{sections/experiments}

\section{Conclusion}
\input{sections/conclusion}




%
\bibliographystyle{IEEEtran}
\bibliography{mybib}


 




\vfill

\end{document}

%% file: sections/abstract.tex

Human pose estimation and tracking are fundamental tasks for understanding human behaviors in videos. 
Existing top-down framework-based methods usually perform three-stage tasks: human detection, pose estimation and tracking. 
Although promising results have been achieved, these methods rely heavily on high-performance detectors and may fail to track persons who are occluded or miss-detected. 
To overcome these problems, in this paper, we develop a novel keypoint confidence network and a tracking pipeline to improve human detection and pose estimation in top-down approaches. Specifically, the keypoint confidence network is designed to determine whether each keypoint is occluded, and it is incorporated into the pose estimation module. 
In the tracking pipeline, we propose the Bbox-revision module to reduce missing detection and the ID-retrieve module to correct lost trajectories, improving the performance of the detection stage. Experimental results show that our approach is universal in human detection and pose estimation, achieving state-of-the-art performance on both PoseTrack 2017 and 2018 datasets.

%% file: sections/introduction.tex
\begin{figure}[t!]
    \centering
    \rotatebox{90}{{~~Baseline}}
    \begin{minipage}[t]{0.31\linewidth}
           \includegraphics[width=\linewidth]{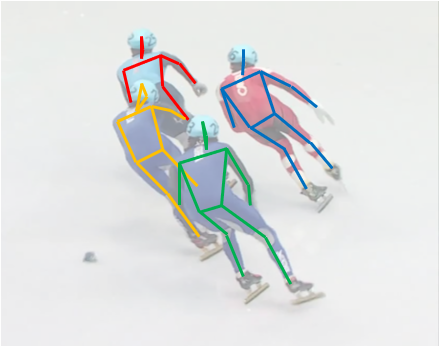}
    \end{minipage}
    \begin{minipage}[t]{0.31\linewidth}
            \includegraphics[width=\linewidth]{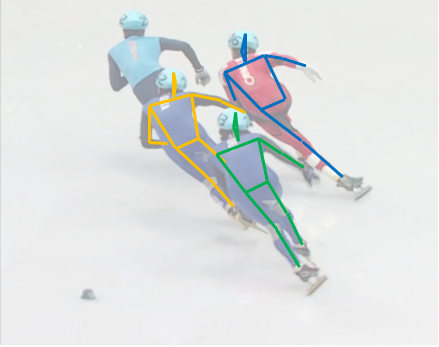}
    \end{minipage}
    \begin{minipage}[t]{0.31\linewidth}
            \includegraphics[width=\linewidth]{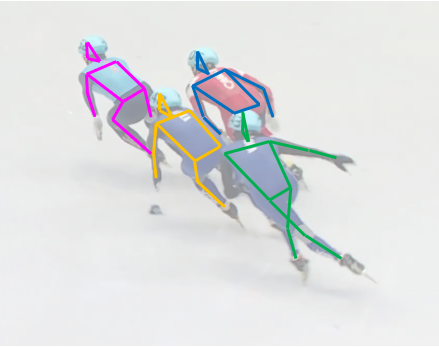}
    \end{minipage}
    
    \rotatebox{90}{{~~~~~~~~Ours}}
    \begin{minipage}[t]{0.31\linewidth}
           \includegraphics[width=\linewidth]{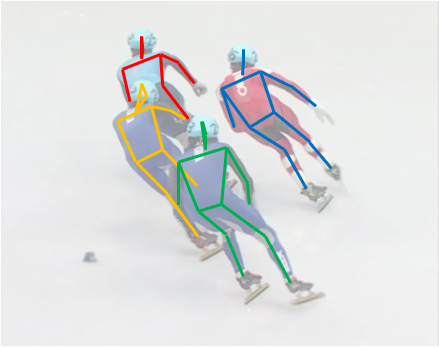}
           \centerline{Frame 31}\medskip
    \end{minipage}
    \begin{minipage}[t]{0.31\linewidth}
            \includegraphics[width=\linewidth]{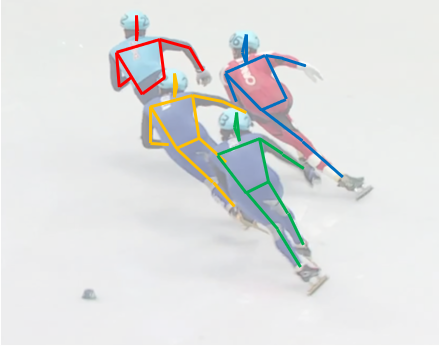}
            \centerline{Frame 33}\medskip
    \end{minipage}
    \begin{minipage}[t]{0.31\linewidth}
            \includegraphics[width=\linewidth]{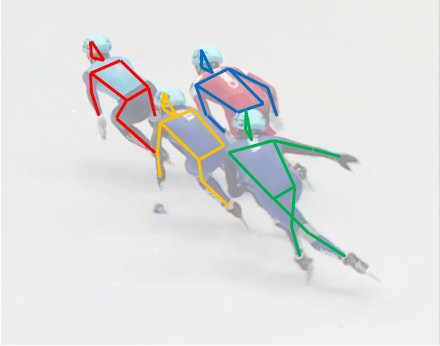}
            \centerline{Frame 40}\medskip
    \end{minipage}
    \caption{The Baseline is a top-down multi-person pose tracking method based on HRNet. The baseline may cause the detector failure in occlusion (Frame 33), which results in loss of tracking trajectory (Frame 40). Although using the same detector as the baseline, our approach overcomes this limitation through the Bbox-revision module in the pose tracking pipeline.}
    \label{fig:fig0}
\end{figure}

Multi-person pose tracking, which intends to detect the body joints of all persons in the video frames and output the pose trajectories over time consistently, is a fundamental task for human-centered video understanding.
It is widely used in human-computer interaction, video surveillance, and action recognition\cite{liu2018recognizing,bao2020pose,8897575,8089370}, etc. 
Multi-person pose tracking tasks serve as integral components in multimedia applications, facilitating advances in domains such as video analysis, gesture recognition, gaming and interactive experiences. These tasks allow precise interpretation and assessment of human movements, thus increasing user interaction, immersion, and comprehensiveness in multimedia contexts.
The multi-person pose tracking can be categorized into bottom-up~\cite{insafutdinov2017arttrack} and top-down~\cite{wang2020combining,zhou2020temporal,yu2018multi,snower202015} methods. 
The former one estimates all joints in each frame and associates each person's joints over time in a spatio-temporal optimization manner without detecting human bounding boxes. 
While the latter pipeline first detects the bounding box of each person and then estimates his/her pose. 
In the final stage, a multi-person tracking methodology is applied to retrieve the trajectories of joints. Thanks to the advancement of object detectors in recent years~\cite{10.1007/978-3-030-58452-8_13, 9932281, 9600874, 9424971, 8445665, zou2019object, 8627998, 9123553, redmon2016you, tan2020efficientdet}, the top-down pipeline has made great progress and has surpassed the bottom-up pipeline to become the mainstream. 
However, there are two barriers that prevent these methods from being perfect: occlusion and fast motion (e.g., Figure~\ref{fig:fig0} ). 
Top-down methods filter keypoints based on heatmaps that are predicted by pose estimators optimized for images instead of video frames.
The estimators suffer from motion blurs, and thus it is hard to produce accurate keypoints. 
Furthermore, occlusions between adjacent persons may fool estimators into making wrong predictions. 
Finally, occlusions and motion blurs may also degrade person detectors, eventually leading to tracking failure. 

\par
In this paper, we attack these two problems with a novel confidence estimation and temporal correction strategy. Specifically, we design a confidence network measuring the visibility of the keypoints in addition to the location probabilities in the heatmap. Then, we build an online tracking pipeline to perform multi-person pose tracking, which consists of three modules, including an association module, an ID-retrieve module, and a Bbox-revision module. Similarly to previous work \cite{girdhar2018detect}, the association module assigns a unique identification number to each pose in frames. We follow \cite{girdhar2018detect} and employ the Hungarian algorithm to achieve this. However, due to challenges such as motion blurs and occlusions, there exist non-matched poses in both sets. We solve this problem using the following two modules. The ID-retrieve module assigns the ID of a person in the current frame who has no matched ID in the previous frame but may be matched in history. The Bbox-revision module serves as an assistant to the detector. 
It helps to repair missing detections by leveraging the motion of persons. 
We also introduce a pose filtering operation to improve the overlapping poses caused by error detections. 
\par
Research conducted by Yang et al.~\cite{yang2021learning} addresses the same issue as our proposed Bbox-revision module, namely the missed detection in the current frame on challenging scenes such as occlusion and fast motion. 
Both work~\cite{yang2021learning} and ours aggregate the pose of the current frame $t$ from the estimator and the predictor.
Work~\cite{yang2021learning} predicts the pose of the current frame $t$ from $n$ historical poses using GNN (Graph Neural Network), while ours predicts the pose of the current frame $t$ from the $t-1$ pose using optical flow. 
Considering that poses in the current frame and historical frames may not be correct, there are two main differences between work~\cite{yang2021learning} and ours.
First, we filter the wrong detections, namely bounding boxes without objects with the score calculated from the proposed keypoint confidence.
Second, we proposed a novel similarity matrix based on our keypoint confidence for one-to-one mapping to remove redundant detections.
Thus, we improve the tracking on missed detection with a novel keypoint confidence. 
Besides, we also propose an ID-retrieve module to correct lost trajectories, which further improves the tracking performance.
\par
To summarize, the main contributions of this study are as follows:
\begin{itemize}
\item We design a keypoint confidence network to measure the visibilities of keypoints, which is helpful to improve the pose estimator's performance on occluded joints; we also propose a tracking pipeline that corrects lost trajectories and miss-detections.
\item The keypoint confidence network and pose tracking pipeline are universal and can be used with different pose estimation networks and human detectors.
\item The proposed method achieves MOTAs of 69.2\%, 72.2\%, and 63.5\% on the 2018 validation set, the 2017 validation set, and the 2017 test set of the PoseTrack dataset, respectively, achieving state-of-the-art performance.
\end{itemize}



%% file: sections/related_work.tex
We briefly review the following two related topics, including multi-person pose estimation and multi-person pose tracking.
\subsection{Multi-person Pose Estimation in Image}
Human pose estimation, which aims to locate the keypoint of the human body in images, can be classified into single-person~\cite{pfister2015flowing,toshev2014deeppose,newell2016stacked,wei2016convolutional} and multi-person pose estimation~\cite{cao2017realtime,fang2017rmpe,song2017thin,papandreou2017towards,pishchulin2016deepcut,sun2019deep}. The multi-person pose estimation is more realistic and challenging. It has received increased attention and made significant progress in recent years~\cite{DBLP:conf/cvpr/TompsonGJLB15,
DBLP:conf/nips/TompsonJLB14,
wei2016convolutional,
DBLP:conf/eccv/BulatT16,
DBLP:conf/cvpr/CarreiraAFM16,
DBLP:conf/cvpr/HuR16,DBLP:conf/cvpr/HuangRSZKFFWSG017,
newell2016stacked,
DBLP:conf/bmvc/RafiLGK16,
xu2022vitpose,
DBLP:conf/fgr/BulatKTP20,
9677941,9858008,8064661}.
Multi-person pose estimation is generally classified into top-down methods~\cite{DBLP:conf/cvpr/PishchulinJATS12,fang2017rmpe,sun2019deep,he2017mask,DBLP:conf/cvpr/ChenWPZYS18,xiao2018simple} 
and bottom-up methods~\cite{DBLP:conf/cvpr/GkioxariHGM14,
DBLP:conf/cvpr/ChenY15,
DBLP:conf/eccv/InsafutdinovPAA16,
pishchulin2016deepcut,
DBLP:conf/eccv/IqbalG16,
DBLP:conf/cvpr/KreissBA19,
DBLP:journals/tits/KreissBA22}.
\par
The two most important components of top-down methods are the human detector and the single-person pose estimator. Top-down methods first detect persons and generate the person bounding boxes. Then, pose estimation is conducted to detect the human pose for each bounding box. Unlike top-down methods, bottom-up methods do not rely on human detectors. 
Bottom-up methods first detect all body joints of every person and then group them by some fitting algorithms to form human poses. 
In this work, we mainly focus on the top-down method. To verify the effectiveness of our method, we will conduct experiments on Hourglass~\cite{newell2016stacked}, SimpleBaselinet~\cite{xiao2018simple} and HRNet~\cite{sun2019deep}. 

\subsection{Multi-person Pose Tracking}

Recently, multi-person pose tracking has received significant attention since the topic was first introduced by the PoseTrack dataset.
Pose estimation in images can be extended to pose tracking in the video by running independently on each frame and then using data association to correctly link the continuous trajectory of each person over time.
Bottom-up methods~\cite{raaj2019efficient,jin2019multi} estimate all joints in each frame and associate the joints in a spatio-temporal optimization manner without detecting human
bounding boxes. For example, Spatio-Temporal Affinity Fields (STAF)~\cite{raaj2019efficient} build upon Part Affinity Fields representation~\cite{cao2017realtime} and propose an architecture that can encode and predict spatio-temporal affinity fields across a video sequence.
\par
Top-down methods~\cite{DBLP:conf/eccv/GuoTLCLW18,
bao2020pose,
yang2021learning,
snower202015} are based on top-down pose estimation and exploiting spatio-temporal context for tracking.
CombDet~\cite{wang2020combining} extended HRNet~\cite{sun2019deep} from 2D to 3D to build a tracking pipeline to alleviate missed detection, but the approach is limited by the size of clip lengths.
TKMRNet~\cite{zhou2020temporal} proposes refinement networks to improve pose precision and design new keypoint similarity metrics in the association module.
Existing top-down methods treat pose estimation and pose tracking as two relatively independent tasks designed and optimized separately. 
Thus, the pose tracking network directly utilizes the estimated pose from the pose estimation network. In this work, we try to boost the tracking performance by leveraging the additional information output from the pose estimation network, i.e., the proposed keypoint confidence.



%% file: sections/methodology.tex
\begin {figure*}[tbh]
\centering
\includegraphics[width=1\textwidth]{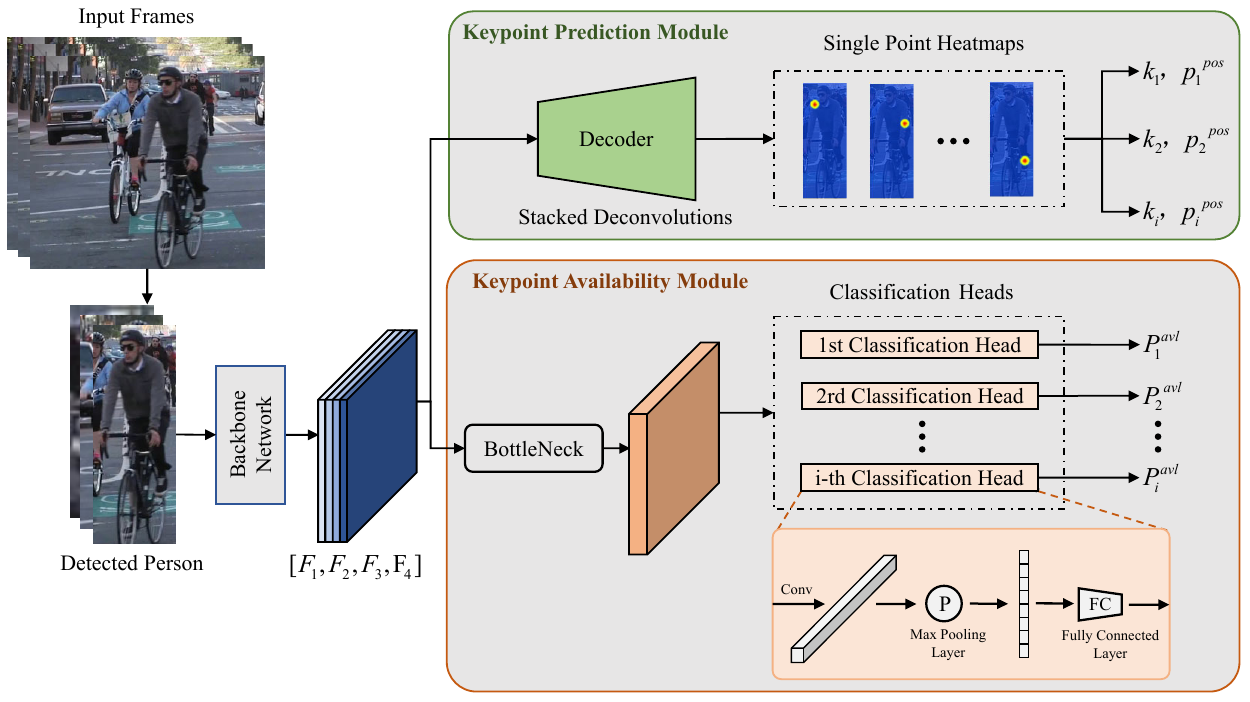}
\caption{
Illustrating the architecture of the Keypoint Confidence Network (KCN). It consists of a keypoint prediction module and a keypoint availability module. $[F_1,F_2,F_3,F_4]$ is the high-resolution representation after multiple-scale fusion with exchange units.
}
\label{fig:kcn}
\end {figure*}
In this section, we introduce the Keypoint Confidence Network (KCN) and the Pose Tracking Pipeline (PTP) for multi-person pose tracking. 
Our method works by first detecting all persons in the current frame and then estimating the keypoints and their confidence for each person by KCN. 
In the tracking stage, we use two modules to correct the tracking: we retrieve the lost trajectories in previous frames using the ID-retrieve module and revise the bounding boxes of persons with the aid of optical flows using the Bbox-revision module. 
In what follows, we introduce each component of the proposed method in detail.
\subsection{Keypoint Confidence Network}
In the process of pose tracking, interactive occlusions can lead to the unavailability of some keypoints, necessitating filtering.
Existing research employs keypoint location probability, namely, the probability of each keypoint at its current location, as the basis for filtering.
However, this approach may lead to several issues: on the one hand, occluded keypoints that are mistakenly assigned to other individuals maintain high location probabilities and are erroneously retained; 
on the other hand, under conditions of frame blurring, location probabilities for each keypoint are generally low, causing inaccurate filtering. 
The occurrence of these errors complicates the setting of keypoint thresholds, undermines keypoint recognition accuracy, and directly impacts tracking performance. 
To address these issues, we exploit the global spatial structure information of the human body for modeling the availability probability of each keypoint. 
And then introduce a Keypoint Confidence Network (KCN). 
By combining keypoint location probability and availability probability, we effectively alleviate the aforementioned issues and subsequently improve tracking performance.
\par
The architecture of our proposed KCN is shown in Figure~\ref{fig:kcn}. 
Our proposed KCN consists of a backbone for extracting features and two parallel branches for pose estimation: the Keypoint Prediction Module (KPM) for predicting keypoint location as well as the location probability and the Keypoint Availability Module (KAM) for estimating the probability of keypoint availability. 
We use HRNet~\cite{sun2019deep} as the backbone for feature extraction.
\par
\textbf{Keypoint Prediction Module.} 
The keypoint prediction module consists of three $3 \times 3$ deconvolution layers and generates $K$ heatmaps, where $K$ is the number of keypoints for each person. 
For each heatmap $M$, we obtain the keypoint location $l$ from the highest response in the heatmap and use its response value as keypoint location probability $p^{loc}$. 
Following~\cite{wang2020combining}, we convert point-wise annotations into heatmaps using 2D Gaussian convolutions and consider them as ground truths of keypoints during training. 
The loss function of this module is defined as

\begin{equation}
L=\frac{1}{K W H} \sum_{k}^{K} \sum_{i}^{W} \sum_{j}^{H}\left\|M_{k i j}-G_{k i j}\right\|_{2}^{2}
\end{equation}
where ${W}$ and ${H}$ represent the width and height of heatmaps. ${M}$ and ${G}$ represent the predicted heatmaps and the ground truth maps.
\par
\textbf{Keypoint Availability Module.} The keypoint availability module is composed of a bottleneck
layer \cite{he2016deep} and $K$ classifier heads to obtain the keypoint availability probability $p^{avl}$. 
The bottleneck layer consists of three convolution layers with sizes of $1 \times 1$, $3 \times 3$, and $1 \times 1$, respectively.
Each layer has a channel size of 384. 
The classifier heads predict the availability probabilities of keypoints. For each head, there are a $1 \times 1$ convolution layer, a global max-pooling layer, and a fully connected layer followed by a softmax layer. The convolution layer has 384 channels as input and outputs 512 channels. 
The fully connected layer has 512 input units and 2 output units.
\par
During training, we use an ${\alpha}$-balanced variant of the focal loss~\cite{lin2017focal} $L_f$ for optimization

\begin{equation}
\mathrm{L}_{f}= \begin{cases}-\alpha(1-p)^{\gamma} \log (p) &, y=1 \\ -(1-\alpha) {p}^{\gamma} \log (1-p) &, y=0\end{cases}
\end{equation}
where $p$ is the keypoint availability probability of $p^{loc}$, and $y$ is the label of keypoint availability. 
We use the default settings where ${\gamma}$ = 2, ${\alpha}$ = 0.25.
\par
\textbf{Keypoint Confidence.} Through KCN, we obtain both keypoint location probability $p^{loc}$ and keypoint availability probability $p^{avl}$. By combining them, we can calculate the keypoint confidence $p^{conf}$ of the $i$-th body joint as follows
\begin{equation}
p_{i}^{conf}=p_{i}^{avl} \times p_{i}^{loc}
\end{equation}

\subsection{Pose Tracking Pipeline}
\begin {figure*}[t!]
\includegraphics[width=1\textwidth]{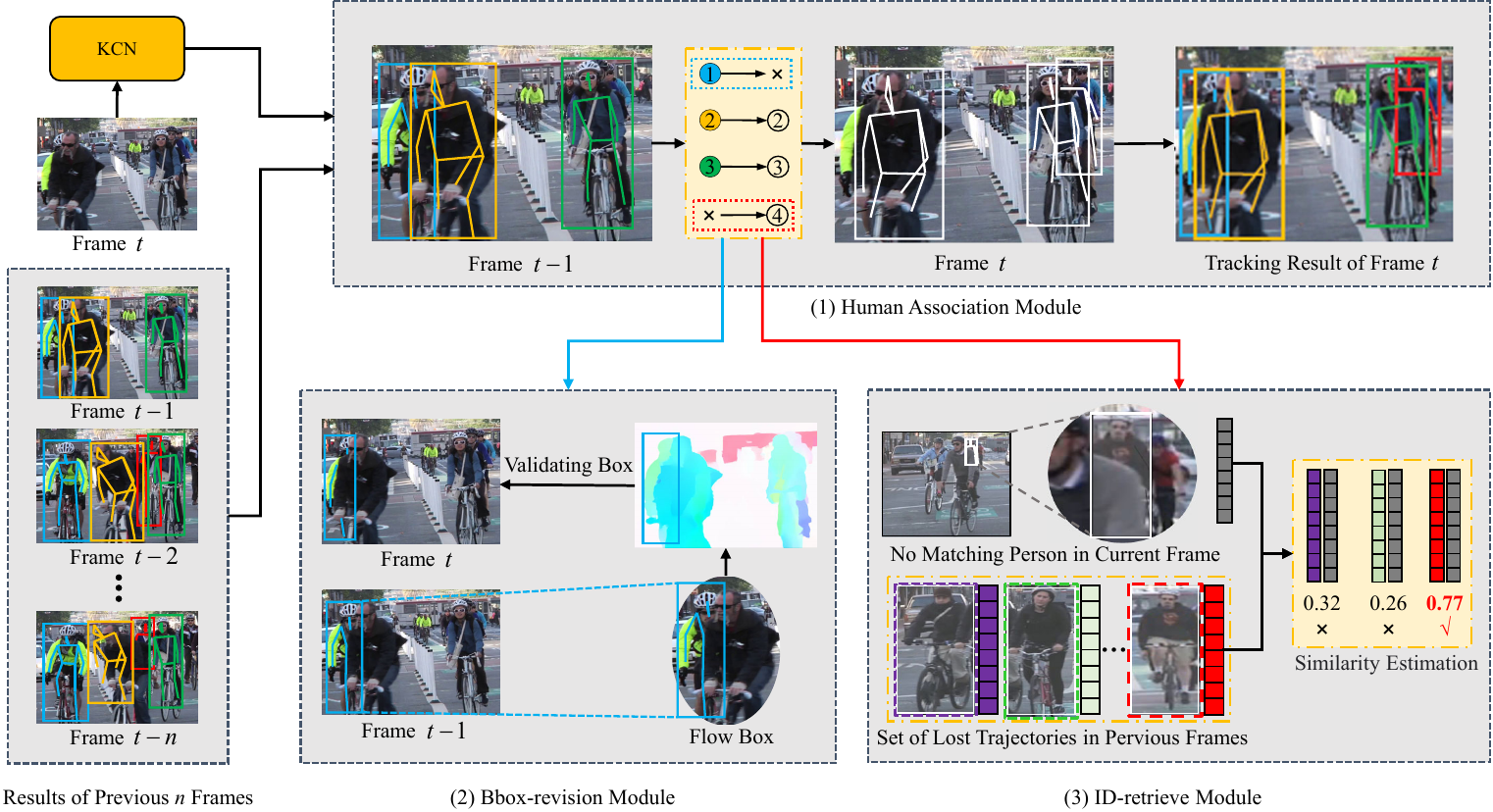}
\caption{Illustration of the proposed pose tracking pipeline. In the first stage, we detect persons and estimate their poses using our keypoint confidence network. Then, in the tracking stage, (1) performs the identity association between frames and (2) generates the bounding boxes from the previous frame for the unmatched trajectories. Finally, (3) identifies a person in the current frame who has no matched ID in the previous frame but may be matched in history.} 
\label{fig:ptp}
\end {figure*}
The illustration of our proposed pose tracking pipeline is shown in Figure~\ref{fig:ptp}.
In the tracking stage, we first use a human association module to match persons. 
Two things may happen:
1) a person is lost in previous frames but shows up again in the current frame; 
2) the detector may miss detecting persons due to occlusion or fast motion. 
We propose the following two modules, the ID-retrieve module and the Bbox-revision module, to solve the above two challenges. 
The ID-retrieve module retrieves lost IDs using a person re-identification technique. 
We use the Bbox-revision module to generate bounding boxes that are missed by the detector in the current frame. The details of each module are given below. 
\par
\textbf{Human Association Module.} 
During tracking, we first analyze the detection and pose estimation results in the current frame and then associate them with the existing trajectories by assigning a unique identification number.
The Human Association Module in our approach addresses the weighted bipartite graph matching problem, aiming to establish an optimal one-to-one correspondence between the trajectory from the previous frame, and the detection and pose estimation results from the current frame.
To accomplish this, we utilized the Hungarian algorithm, which requires the construction of a weight matrix. Specifically, to measure the similarity of a person across frames, we employ the Object Keypoint Similarity (OKS)~\cite{oks}, an evaluation metric in multi-person pose tracking commonly used in the COCO keypoints challenge and PoseTrack challenge. 
Consider both pose similarity and distance, ensuring only poses that are similar and close poses receive higher weights. 
\textbf{Bbox-revision Module.} 
The Bbox-revision module is used to reduce missing detection in the current frame. 
As shown in Figure~\ref{fig:ptp} (2), the person detected in the previous frame failed to be detected in the current frame, resulting in the unmatched trajectory. 
To address this issue, we first use the optical flow sub-module to generate bounding boxes from the previous frame and verify their correctness by a pose filtering sub-module. 
The verified bounding boxes are reserved and identified with the ID of the corresponding unmatched trajectories.
\par
For the optical flow sub-module, we apply the RAFT (Recurrent All-Pairs Field Transforms) \cite{teed2020raft} method to get the offset of each pixel in the previous frame. 
Thus, the position of each pixel in the unmatched pose trajectory in the current frame is provided and the minimum bounding rectangle is used as the bounding box.
\par
After bounding boxes are generated, the pose filtering sub-module is used to verify their correctness in two steps. 
First, KCN is applied to estimate human poses, and the average keypoint confidence is used as the score of each bounding box for filtering. 
Then, we propose a novel similarity metric that evaluates the overlap between the bounding boxes generated by optical flow and those detected in the current frame. 
This approach addresses the challenge of sub-region overlapping of bounding boxes generated by optical flow. 
The optical flow sub-module may produce wrong trajectories due to the disappearance of the person or missing detection in the previous stage. 
Therefore, in the scene where the human is interlaced and occluded, the Non-Maximum Suppression (NMS) algorithm based on the Intersection over Union (IoU) and OKS cannot filter out the overlapping box, which ultimately affects the final performance. 
To address this issue, we propose a new similarity metric that calculates the overlap of poses by focusing on the shared keypoint regions within both boxes, allowing a more efficient filtering process for overlapping bounding boxes. 
It is defined as 
\begin{equation}
\delta\left(I o U_{p, q}>0.1\right) \frac{\sum_{i} e^{-\frac{d_{i}^{2}}{2 s^{2} k_{i}^{2}}} \delta\left(p_{i}^{\operatorname{conf}}>\theta\right) \delta\left(q_{i}^{\operatorname{conf}}>\theta\right)}{\sum_{i} \delta\left(p_{i}^{\operatorname{conf}}>\theta\right) \delta\left(q_{i}^{\operatorname{conf}}>\theta\right)}
\end{equation}
where the $d_i^2$ is the euclidean distance between person $p$ and person $q$. 
$\delta$ is an indicator function. 
$q_i^{conf}$ and $p_i^{conf}$ are the $i$-th keypoint confidences of person $p$ and person $q$. $s$ is the object scale, and $k_i$ is a per-keypoint constant that controls falloff. $\theta$ is the confidence threshold. 
$IoU_{p, q}$ computes the IoU metric between person $p$ and person $q$, dealing with special cases where two persons do not overlap.
\par
\textbf{ID-retrieve Module.} 
The OKS-based human association may fail when occlusion and persons walk out of sight.
As shown in Figure~\ref{fig:ptp} (3), to remedy such matching failures, we propose the ID-retrieve module to improve the robustness of pose tracking. 
\par
For the input of this module, we reuse the features of the KCN backbone to perform matching followed by adaptive average pooling operations to obtain compact pedestrian features. 
Meanwhile, we maintain a feature set for lost persons. 
For a new detection in the current frame, we first perform feature matching with historical persons.
If the similarity score reaches the threshold, we assign its historical person ID; otherwise, we assign a new ID.

%% file: sections/experiments.tex
\begin{figure}[tb]
    \centering
    \includegraphics[width=0.55\linewidth]{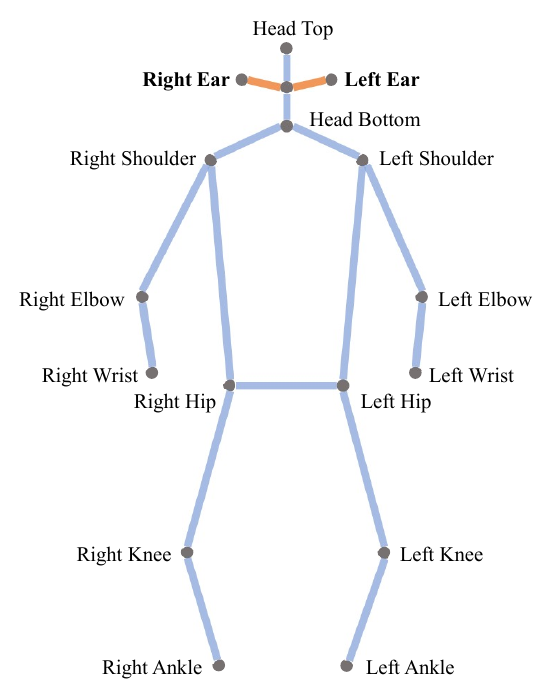}
    \caption{Illustration of body joints in PoseTrack 2017 and PoseTrack 2018 datasets.}
    \label{fig:joints_instruction}
\end{figure}
\par
\begin{table}
\centering
\caption{The description of PoseTrack 2017 Dataset.}
\setlength{\tabcolsep}{2.7mm}{
\begin{tabular}{l|llll}
\toprule
               & Num Poses~    & Num Trajectories~ & Num Videos \\
\midrule
Training set   & 61178           & 2437               & 292          \\
Validation set & 18996           & 695                & 50           \\
Test set       & 73471           & 2334               & 208          \\
Total          & \textbf{153615} & \textbf{5446}      & \textbf{550} \\
\bottomrule
\end{tabular}}
\label{tab:pt2017}
\end{table}
\subsection{Datasets and Evaluation Metrics}
We evaluate the proposed method on PoseTrack, which is a large-scale benchmark for multi-person pose estimation and pose tracking in videos. 
It contains several video sequences with a lot of annotated poses with various activities being performed. 
PoseTrack has the 2017 and 2018 versions of this benchmark. Each dataset has a publicly available training set and validation set, as well as an evaluation server for benchmarking on a held-out test set.
PoseTrack 2017 annotates 15 body parts for each body pose, including the head, nose, neck, shoulders, elbows, wrists, hips, knees, and ankles; while PoseTrack 2018 annotates two more ears, as shown in Figure~\ref{fig:joints_instruction}.
For our experiments, only the original 15 body joints are used for both training and inference.
We conduct experiments on both PoseTrack 2017 and PoseTrack 2018.
Especially, PoseTrack 2017 includes 250 videos for training, 50 videos for validation, and 214 videos for tests, as shown in Table~\ref{tab:pt2017}.
PoseTrack 2018 is expanded on the basis of PoseTrack2017, including 593 videos for training, 170 videos for validation, and 375 videos for the test.
This is more than double the amount of data from PoseTrack 2017.
In the training set, the videos are densely annotated 30 frames from the center of frames. 
In the validation and testing set, the videos are densely annotated 30 frames of the middle, and afterward annotated every fourth frame. In addition, we use the COCO dataset to pre-train the multi-person pose estimation model used in our experiments.
\par
\begin{table}[tbh]
\centering
\caption{The Pose Tracking performance in MOTA~(\%) with the different confidence threshold on PoseTrack validation datasets.}
\label{tab:abs_keypoint_threshold}
\setlength{\tabcolsep}{1.7mm}{
\begin{tabular}{c|l|lllllll} 
\toprule
\multirow{2}{*}{Dataset}        & \multirow{2}{*}{Method} & \multicolumn{7}{c}{Threshold Value}             \\ 
\cline{3-9}
                                &                         & 0.25 & 0.30 & 0.35 & 0.40 & 0.45 & 0.50 & 0.55  \\ 
\midrule
\multirow{2}{*}{\begin{tabular}[c]{@{}c@{}}PoseTrack\\2017\end{tabular}} & Baseline                & 61.2 & 63.0 & 64.5 & 65.6 & 66.2 & 66.5 & 65.3  \\
                                & Ours                    & 69.2 & 69.5 & 69.5 & 69.0 & 68.3 & 67.1 & 65.4  \\ 
\midrule
\multirow{2}{*}{\begin{tabular}[c]{@{}c@{}}PoseTrack\\2018\end{tabular}} & Baseline                & 55.4 & 57.7 & 59.6 & 61.3 & 62.5 & 63.1 & 62.5  \\
                                & Ours                    & 65.9 & 66.5 & 66.6 & 66.1 & 65.6 & 64.8 & 63.1  \\
\bottomrule
\end{tabular}}
\end{table}

\begin{figure}[tb]
    \centering
    \includegraphics[width=1\linewidth]{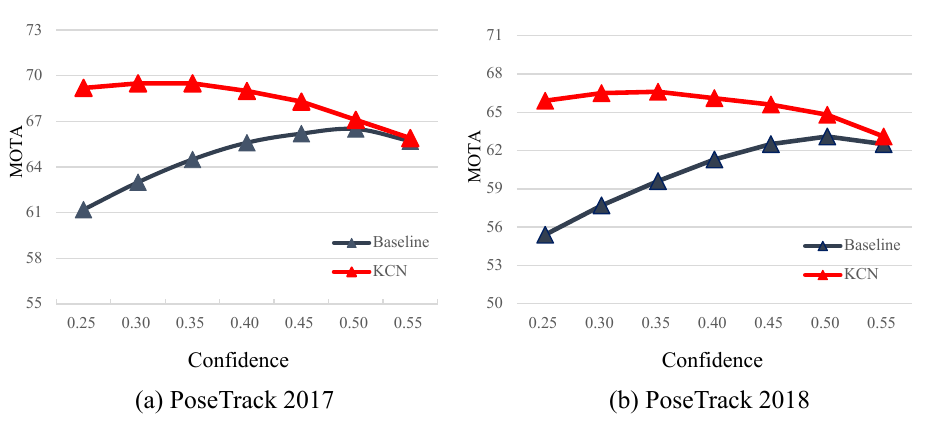}
\caption{
The impact of the different confidence thresholds on PoseTrack validation datasets. The numbers in the figure refer to MOTA~(\%).
}
    \label{fig:abs_cfd_thresholds}
\end{figure}
\par
\begin{table}
\centering
\caption{Effectiveness analysis of KCN by pose tracking performance (MOTA) on PoseTrack validation datasets.}
    \label{tab:abs_KCN_effectiveness1}
\setlength{\tabcolsep}{0.8mm}{
\begin{tabular}{c|l|ccccccc|l} 
\toprule
Dataset                         & Method   & Head & Shou & Elb  & Wri  & Hip  & Knee & Ankl & Total                \\ 
\midrule
\multirow{2}{*}{\begin{tabular}[c]{@{}c@{}}PoseTrack\\2017\end{tabular}} & Baseline & 73.6 & 76.4 & 61.1 & 55   & 63.6 & 64.8 & 52.7 & 64.5                 \\
                                & Ours     & 76.5 & 77.8 & 70.0   & 63.2 & 66.2 & 68.9 & 60.5 & 69.5(+\textbf{5.0})  \\ 
\midrule
\multirow{2}{*}{\begin{tabular}[c]{@{}c@{}}PoseTrack\\2018\end{tabular}} & Baseline & 68.0   & 71.7 & 55.7 & 51.6 & 58.5 & 58.2 & 49.4 & 59.6                 \\
                                & Ours     & 69.8 & 74.7 & 69.7 & 62.3 & 63.9 & 64.7 & 59.2 & 66.6(+\textbf{7.0})  \\
\bottomrule
\end{tabular}}
\end{table}

\begin{table}[tbh]
\centering
\caption{
Keypoint statistical analysis of wrong detection and missing detection performed on the validation set of PoseTrack 2018.}
    \label{tab:abs_KCN_effectiveness2}
\setlength{\tabcolsep}{.7mm}{
\begin{tabular}{c|l|ll|l} 
\toprule
Dataset                         & Method   & Wrong Detection         & Missing Detection      & Total                    \\ 
\midrule
\multirow{2}{*}{\begin{tabular}[c]{@{}c@{}}PoseTrack\\2017\end{tabular}} & Baseline & 22249                   & 42099                  & 64348                    \\
                                & Ours     & 19940(-\textbf{10.4\%}) & 38270(-\textbf{9.1\%}) & 58210(-\textbf{9.5\%})   \\ 
\midrule
\multirow{2}{*}{\begin{tabular}[c]{@{}c@{}}PoseTrack\\2018\end{tabular}} & Baseline & 74353                   & 97297                  & 171650                   \\
                                & Ours     & 66058(-\textbf{11.6\%}) & 88565(-\textbf{9.0\%}) & 154623(-\textbf{9.9\%})  \\
\bottomrule
\end{tabular}}
\end{table}

\par
\begin{table*}[tb]
\centering
\caption{Generalization of KCN across three top-down multi-person pose estimation networks by pose tracking performance in MOTA~(\%). }
\label{tab:abs_KCN_generalization}
\setlength{\tabcolsep}{4mm}{
\begin{tabular}{l|l|c|lllllll|l} 
\toprule
Dataset        & Method       & KCN & Head & Shou & Elb  & Wri  & Hip  & Knee & Ankl & Total                \\ 
\midrule
\multirow{6}{*}{PoseTrack 2017} & Hourglass    &     & 71.6 & 71.8 & 57.6 & 49.2 & 52.6 & 54.2 & 43.5 & 58.2                 \\
               & Hourglass    & \Checkmark   & 73.8 & 73.3 & 62.6 & 53.1 & 54.2 & 55.6 & 49.9 & 61.2(\textbf{+3.0})  \\
               & Pose\_ResNet &     & 72.5 & 71.8 & 57.7 & 48.4 & 58.7 & 60.3 & 49.2 & 60.6                 \\
               & Pose\_ResNet & \Checkmark   & 73.5 & 72.6 & 63.4 & 53.1 & 59.3 & 62.6 & 55.5 & 63.6(\textbf{+3.0})  \\
               & HRNet        &     & 73.6 & 76.4 & 61.1 & 55.0   & 63.6 & 64.8 & 52.7 & 64.5                 \\
               & HRNet        & \Checkmark   & 76.5 & 77.8 & 70.0   & 63.2 & 66.2 & 68.9 & 60.5 & 69.5(\textbf{+5.0})  \\ 
\midrule
\multirow{6}{*}{PoseTrack 2018} & Hourglass    &     & 65.6 & 66.5 & 52.6 & 45.5 & 50.9 & 51.0   & 42.4 & 54.3                 \\
               & Hourglass    & \Checkmark   & 67.7 & 68.0   & 58.9 & 50.1 & 53.3 & 53.6 & 48.5 & 57.9(\textbf{+3.6})  \\
               & Pose\_ResNet &     & 66.4 & 68.2 & 52.4 & 46.1 & 56.3 & 55.5 & 47.0   & 56.7                 \\
               & Pose\_ResNet & \Checkmark   & 67.9 & 69.3 & 61.6 & 52.3 & 59.2 & 59.8 & 54.4 & 61.1(\textbf{+4.4})  \\
               & HRNet        &     & 68.0   & 71.7 & 55.7 & 51.6 & 58.5 & 58.2 & 49.4 & 59.6                 \\
               & HRNet        & \Checkmark  & 69.8 & 74.7 & 69.7 & 62.3 & 63.9 & 64.7 & 59.2 & 66.6(\textbf{+7.0})  \\
\bottomrule
\end{tabular}}
\end{table*}

The performance of our method is evaluated from two aspects. We evaluate the accuracy of multi-person pose estimation using the mean Average Precision~(mAP).
We use Multiple Object Tracking Accuracy~(MOTA) to evaluate the performance of trackers at keeping trajectories, including the performance of false positives, misses, and mismatches. MOTA is the main evaluation metric for PoseTrack Benchmark, which is defined as
\begin{equation}
    \mathrm{MOTA}=1-\frac{\sum_{t}\left(F P_{t}+F N_{t}+I D S_{t}\right)}{\sum_{t} g_{t}}
\end{equation}
where the subscript $t$ refers that current values are computed at the $t$-th frame. 
$FP_t$, $FN_t$, and $IDS_t$ denote the number of false positives, missed targets, and identity switches, respectively, at time $t$.
$g_t$ stands for ground truth.

\par

\subsection{Implementation Details}
In the human detection stage, we use an HTC detector~\cite{chen2019hybrid} to detect all the person instances in frames and extract crops of size 384×288 around detected person instances as input to our proposed keypoint confidence network. 
We use the pre-trained model trained on the COCO dataset in mmdetection~\cite{mmdetection}. 
Note that no additional fine-tuning of the detectors on the PoseTrack dataset is performed. 
In Non-Maximum Suppresion~(NMS) operation for human detection, we changed the metric from IoU to OKS metric and set the threshold to 0.6.
\par
In the pose estimation stage, our keypoint confidence network used HRNet~\cite{sun2019deep} as the backbone.
In the training phase, we first train the keypoint prediction module and then attach the point confidence module to refine the two modules together.
Different from other methods, we combine the training set and the validation set of COCO dataset and the training set of the PoseTrack dataset for model training. 
The COCO dataset and the PoseTrack dataset do not agree on two joint types.
The PoseTrack dataset contains the Head top and Head bottom joint points, while the COCO dataset contains the left ear and right ear joint points. 
Since the two different types of joint points between these two datasets are similar in position and both belong to the joint points of the face, we directly use the visibility of the ear instead of the visibility of the head for training.
In particular, we use the default parameters of the HRNet in the pose estimation task for training. 
When training the point confidence module, we trained a total of 20 epochs to fine-tune the model.
\par

To train the ID-retrieve module of the proposed pose tracking pipeline, we built a dataset based on PoseTrack 2018, which contains 119656 images with 4613 person tags. 
We use the Euclidean distance to measure the similarity between two features. 
When the similarity is less than the threshold value~(e.g. 100), we consider it to be a person.
\par
\subsection{Ablation Study}
In this section, we provide ablation experiments and analysis to demonstrate the effectiveness of our proposed framework and the strength of the key components. 
For clarity, we denote the Keypoint Confidence Network as KCN and the Pose Tracking Pipeline as PTP. 


\textit{\textbf{Analysis of Confidence Threshold.}}
For top-down methods, the keypoint confidence threshold significantly impacts the final permanence. 
For a fair comparison, our baseline network only replaces the HRNet-based keypoint confidence network with an HRNet-based pose estimation method. 
As shown in Table~~\ref{tab:abs_keypoint_threshold}, with the confidence threshold ranging from 0.25 to 0.55, we compare the multi-person pose tracking performance variations of our KCN and the baselines on PoseTrack 2017 and PoseTrack 2018, respectively. 
We also visualize the result in Figure~\ref{fig:abs_cfd_thresholds}. 
As can be observed, the performance variation curves of KCN are much flatter between the two, implying that our KCN reduces the impact of the confidence threshold compared to our baseline. 
Meanwhile, our KCN outperforms the baseline method stably as the threshold changes. 
We observe that the performance of KCN begins to decline when the confidence threshold is greater than 0.35. 
It may be caused by the mistaken filtration of correct keypoints. 
The confidence threshold is set to 0.35 for the rest of the experiments.
\par

\begin{table}[tb]
\centering
 \caption{Generalization of the proposed pose tracking pipeline with different backbones on PoseTrack 2018 validation set. Performance is evaluated by mAP~(\%) and MOTA~(\%) for pose estimation and pose tracking in the multi-person pose estimation and tracking task, respectively.}
  \label{tab:abs_ptp}
\setlength{\tabcolsep}{1.9mm}{
\begin{tabular}{c|c|c|l|l} 
\toprule
Detector              & ID-retrieve & Bbox-revision & mAP & MOTA \\ 
\midrule
\multirow{4}{*}{\footnotesize{HTC~\cite{chen2019hybrid}}}  &       &       & 76.6 & 66.6  \\
                      &\Checkmark       &       & 76.6 & 67.3(\textbf{+0.7})  \\
                      &       &\Checkmark       & 78.1(\textbf{+1.5}) & 68.5(\textbf{+1.9})  \\
                      &\Checkmark       &\Checkmark       & 78.1(\textbf{+1.5}) & 69.2(\textbf{+2.6})  \\
\cline{2-5}
\midrule
\multirow{4}{*}{\footnotesize{YOLOv5~\cite{glenn_jocher_2020_4154370}}} &       &       & 76.3 & 66.9  \\
                      &\Checkmark       &       & 76.3 & 67.7(\textbf{+0.8})  \\
                      &       &\Checkmark       & 77.1(\textbf{+0.8}) & 68.0(\textbf{+1.1})  \\
                      &\Checkmark       &\Checkmark       & 77.1(\textbf{+0.8}) & 68.8(\textbf{+1.9})  \\ 
\cline{2-5}
\bottomrule

\end{tabular}}
\end{table}

\begin{table}
\centering
\caption{Effectiveness analysis of each component in Pose Tracking Pipeline by pose estimation performance in mAP~(\%) and pose tracking performance in MOTA~(\%) in multi-person pose estimation and tracking task on the validation set of PoseTrack datasets.}
\label{tab:abs_ptp_effective}
\setlength{\tabcolsep}{1.7mm}{
\begin{tabular}{l|cc|l|l} 
\toprule
Dataset                         & ID-retrieve & Bbox-revision & mAP                 & MOTA                 \\ 
\midrule
\multirow{4}{*}{PoseTrack 2017} &             &               & 80.0                  & 64.5                 \\
                                &      \Checkmark       &               & 80.0                  & 65.4(+\textbf{0.9})  \\
                                &             &     \Checkmark          & 81.8(+\textbf{1.8}) & 67.4(+\textbf{2.9})  \\
                                &      \Checkmark       &       \Checkmark        & 81.8(+\textbf{1.8}) & 67.8(+\textbf{3.3})  \\ 
\midrule
\multirow{4}{*}{PoseTrack 2018} &             &               & 78.3                & 59.6                 \\
                                &      \Checkmark       &               & 78.3                & 60.5(+\textbf{0.9})  \\
                                &             &      \Checkmark         & 80.1(+\textbf{1.8}) & 62.4(+\textbf{2.8})  \\
                                &       \Checkmark      &        \Checkmark       & 80.1(+\textbf{1.8}) & 62.7(+\textbf{3.1})  \\
\bottomrule
\end{tabular}}
\end{table}

\par

\begin{table*}[tb]
\centering
\caption{Ablation study of components KCN and PTP on pose tracking. Trackers are evaluated by MOTA metric~(\%).}
\label{tab:abs_kcn_ptp}
\setlength{\tabcolsep}{4.3mm}{
\begin{tabular}{l|cc|ccccccc|l} 
\toprule
Dataset                         & KCN & PTP & Head & Shou & Elb  & Wri  & Hip  & Knee & Ankl & Total                \\ 
\midrule
\multirow{4}{*}{PoseTrack 2017} &     &     & 73.6 & 76.4 & 61.1 & 55   & 63.6 & 64.8 & 52.7 & 64.5                 \\
                                & \Checkmark   &     & 76.5 & 77.8 & 70.0   & 63.2 & 66.2 & 68.9 & 60.5 & 69.5(+\textbf{5.0})  \\
                                &     & \Checkmark   & 76.7 & 79.7 & 64.8 & 59.4 & 67.0   & 67.3 & 55.5 & 67.8(+\textbf{3.4})  \\
                                & \Checkmark   & \Checkmark   & 79.5 & 81.2 & 72.8 & 66   & 69.5 & 70.8 & 61.8 & 72.2(+\textbf{7.7})  \\ 
\midrule
\multirow{4}{*}{PoseTrack 2018} &     &     & 68.0   & 71.7 & 55.7 & 51.6 & 58.5 & 58.2 & 49.4 & 59.6                 \\
                                & \Checkmark   &     & 69.8 & 74.7 & 69.7 & 62.3 & 63.9 & 64.7 & 59.2 & 66.6(+\textbf{7.0})  \\
                                &     & \Checkmark   & 70.3 & 74.5 & 59.6 & 55.8 & 61.3 & 61.2 & 52.6 & 62.7(+\textbf{3.1})  \\
                                & \Checkmark   & \Checkmark   & 72.2 & 77.4 & 72.4 & 64.7 & 66.1 & 67.2 & 61.3 & 69.2(+\textbf{9.6})  \\
\bottomrule
\end{tabular}}
\end{table*}

\begin{table*}[tbh]
\centering
\caption{Comparison with state-of-the-art methods on multi-person pose estimation (with keypoint filtering) on the Validation sets of PoseTrack dataset in terms of mAP metric~(\%). '-' indicates that the result is not provided in the referred paper.}
\label{tab:posetrack_kpf_vs}
\setlength{\tabcolsep}{4.5mm}{
\begin{tabular}{l|l|ccccccc|c}
\toprule
Dataset        & Method                                                                                 & Head & Shou & Elb  & Wri  & Hip  & Knee & Ankl & Total  \\ 
\midrule
\multirow{11}{*}{PoseTrack 2017} & BUTD~\cite{jin2017towards}       & 79.1 & 77.3 & 69.9 & 58.3 & 66.2 & 63.5 & 54.9 & 67.8   \\
               & RPAF~\cite{zhu2017multi}       & 83.8 & 84.9 & 76.2 & 64   & 72.2 & 64.5 & 56.6 & 72.6   \\
               & ArtTrack~\cite{insafutdinov2017arttrack}   & 78.7 & 76.2 & 70.4 & 62.3 & 68.1 & 66.7 & 58.4 & 68.7   \\
               & PoseFlow~\cite{xiu2018pose}   & 66.7 & 73.3 & 68.3 & 61.1 & 67.5 & 67.0   & 61.3 & 66.5   \\
               & STAF~\cite{raaj2019efficient}       & -    & -    & -    & 65.0   & -    & -    & 62.7 & 72.6   \\
               & ST-Embed~\cite{jin2019multi}   & 83.8 & 81.6 & 77.1 & 70.0   & 77.4 & 74.5 & 70.8 & 77     \\
               & DAT~\cite{girdhar2018detect}        & 67.5 & 70.2 & 62.0   & 51.7 & 60.7 & 58.7 & 49.8 & 60.6   \\
               & FlowTrack~\cite{xiao2018simple}   & 81.7 & 83.4 & 80.0   & 72.4 & 75.3 & 74.8 & 67.1 & 76.9   \\
               & TKMRNet~\cite{zhou2020temporal}    & 85.3 & 88.2 & 79.5 & 71.6 & 76.9 & 76.9 & \textbf{73.1} & 79.5   \\
               & LDGNNTrack~\cite{yang2021learning} & \textbf{88.4} & \textbf{88.4} & \textbf{82.0}   & 74.5 & \textbf{79.1} & 78.3 & \textbf{73.1} & \textbf{81.1}   \\
               & Ours                                                                                   & 86.6 & 87   & 80.1 & \textbf{75.5} & 77.3 & \textbf{78.6} & 71.6 & 80.0     \\ 
\midrule
\multirow{7}{*}{PoseTrack 2018} & STAF~\cite{raaj2019efficient}       & -   & -    & -    & 64.7 & -    & -    & 62   & 70.4   \\
               & TML++~\cite{hwang2019pose}      & -    & -    & -    & -    & -    & -    & -    & 74.6   \\
               & TKMRNet~\cite{zhou2020temporal}    & -    & -    & -    & -    & -    & -    & -    & 76.7   \\
               & LightTrack~\cite{ning2020lighttrack}  & -    & -    & -    & -    & -    & -    & -    & 77.3   \\
               & LDGNNTrack~\cite{yang2021learning} & 80.6 & 84.5 & 80.6 & 74.4 & 75.0   & 76.7 & 71.9 & 77.9   \\
               & SKCTrack~\cite{rafi2020self}   & -    & -    & -    & -    & -    & -    & -    & \textbf{79.2}   \\
               & Ours                                                                                   & 80.6 & 85.3 & 80.7 & 74.3 & 76.1 & 76.7 & 71.9 & 78.1   \\
\bottomrule
\end{tabular}}
\end{table*}

\begin{table*}[tb]
\centering

\caption{Comparison with state-of-the-art methods on pure multi-person pose estimation (without keypoint filtering) on the validation sets of PoseTrack dataset in terms of mAP~(\%).}
\label{tab:posetrack_pe_vset}
\setlength{\tabcolsep}{4.5mm}{
\begin{tabular}{l|l|ccccccc|c}
\toprule
Dataset        & Method                                                                                 & Head          & Shou          & Elb           & Wri           & Hip           & Knee          & Ankl          & Total          \\ 
\midrule
\multirow{5}{*}{PoseTrack 2017} & SKCTrack~\cite{rafi2020self}   & 86.1          & 87.0          & 83.4          & 76.4          & 77.3          & 79.2          & 73.3          & 80.8           \\
               & PoseWarper~\cite{bertasius2019learning} & 81.4          & 88.3          & 83.9          & 78.0          & 82.4          & 80.5          & 73.6          & 81.2           \\
               & CombDet~\cite{wang2020combining}    & 89.4          & 89.7          & 85.5          & 79.5          & 82.4          & 80.8          & 76.4          & 83.8           \\
               & LDGNNTrack~\cite{yang2021learning} & \textbf{90.9} & 90.7          & 86.0          & 79.2          & \textbf{83.8} & \textbf{82.7} & \textbf{78.0} & \textbf{84.9}  \\
               & Ours                                                                                   & 89.5          & \textbf{90.9} & \textbf{87.6} & \textbf{81.8} & 81.1          & 82.6          & 76.1          & 84.6           \\ 
\midrule
\multirow{5}{*}{PoseTrack 2018} & SKCTrack~\cite{rafi2020self}   & \textbf{86.0} & 78.3          & 84.8          & 78.3          & 79.1          & 81.1          & 75.6          & 82.0           \\
               & KeyTrack~\cite{snower202015}   & 84.1          & 87.2          & 85.3          & 79.2          & 77.1          & 80.6          & 76.5          & 81.6           \\
               & CombDet~\cite{wang2020combining}    & 84.9          & 87.4          & 84.8          & 79.2          & 77.6          & 79.7          & 75.3          & 81.5           \\
               & LDGNNTrack~\cite{yang2021learning} & 85.1          & 87.7          & 85.3          & 80.0          & \textbf{81.1} & \textbf{81.6} & \textbf{77.2} & 82.7           \\
               & Ours                                                                                   & 85.1          & \textbf{88.9} & \textbf{86.4} & \textbf{80.7} & 80.9          & 81.5          & 77.0          & \textbf{83.1}  \\
\bottomrule
\end{tabular}}
\end{table*}
\textit{\textbf{Analysis of Keypoint Confidence Network.}}
In this part, we first evaluate the effectiveness of the proposed Keypoint Confidence Network (KCN) on the validation set of PoseTrack datasets.
Our baseline network is the same as in the previous experiment.
As shown in Table~\ref{tab:abs_KCN_effectiveness1}, we compare our KCN with the baseline on the multi-person pose estimation and tracking task, where the performance is evaluated as MOTA and all joints are counted.
As can be seen, after applying KCN, the overall MOTA metrics improve significantly by 5.0\% and 7.0\% on PoseTrack 2017 and PoseTrack 2018, respectively. 
Besides, our KCN outperforms the baseline model on all joints, especially for joints at the elbow and ankle on PoseTrack 2018 dataset, where the MOTA metrics improved by 14.0\% and 9.8\%, respectively. 
We also observed that on more challenging tracking areas, such as elbows, ankles, and wrists, the performance gets impressive improvement with our KCN.
To further demonstrate the effectiveness of KCN, we count the keypoint numbers of wrong detection and missing detection on PoseTrack 2018 dataset, as shown in Table~\ref{tab:abs_KCN_effectiveness2}.
Compared to the baseline method, our KCN shows an 11.6\% reduction in false detection and a 9.0\% reduction in missed detection, for a total reduction of 9.9\%. 
We believe the reason is that the baseline method only uses location probability, which may cause failure when filtering keypoints. 
For example, obscured keypoints may get high location probabilities as they are incorrectly labeled to other persons, thus they will be incorrectly detected; in the case of frame blurring, keypoints will get low location probabilities, thus they will be incorrectly filtered, resulting in missed detection.
\par
We also verify the generalisability of KCN with three different top-down multi-person pose estimation networks, namely HRNet~\cite{sun2019deep}, Hourglass~\cite{newell2016stacked} and Pose$\_$ResNet~\cite{xiao2018simple} on both PoseTrack 2017 and PoseTrack 2018 validation sets, as shown in Table~\ref{tab:abs_KCN_generalization}.
The experimental results indicate that our method can stably improve the performance with different pose estimation networks and has strong generality. 
Our KCN can be plugged into top-down framework-based multi-person pose tracking methods to improve the performance of multi-person pose tracking.
\par

\begin{table*}[tb]
\centering
\caption{Comparison with state-of-the-art methods on multi-person pose tracking on the Validation sets of PoseTrack dataset in terms of MOTA~(\%). '-' indicates that the result is not provided in the referred paper.}
\label{tab:posetrack_pt_vs}
\setlength{\tabcolsep}{4.5mm}{
\begin{tabular}{l|l|ccccccc|c} 
\toprule
Dataset        & Method                                                                                 & Head & Shou & Elb  & Wri  & Hip  & Knee & Ankl & Total  \\ 
\midrule
\multirow{10}{*}{PoseTrack 2017} & BUTD~\cite{jin2017towards}       & 71.5 & 70.3 & 56.3 & 45.1 & 55.5 & 50.8 & 37.5 & 56.4   \\
               & ST-Embed~\cite{jin2019multi}   & 78.7 & 79.2 & 71.2 & 61.1 & 74.5 & 69.7 & 64.5 & 71.8   \\
               & DAT~\cite{girdhar2018detect}        & 61.7 & 65.5 & 57.3 & 45.7 & 54.3 & 53.1 & 45.7 & 55.2   \\
               & FlowTrack~\cite{xiao2018simple}   & 73.9 & 75.9 & 63.7 & 56.1 & 65.5 & 65.1 & 53.5 & 65.4   \\
               & PGPT~\cite{bao2020pose}       & -    & -    & -    & -    & -    & -    & -    & 67.1   \\
               & SKCTrack~\cite{rafi2020self}   & -    & -    & -    & -    & -    & -    & -    & 68.3   \\
               & CombDet~\cite{wang2020combining}    & 80.5 & 80.9 & 71.6 & 63.8 & 70.1 & 68.2 & 62.0 & 71.6   \\
               & TKMRNet~\cite{zhou2020temporal}    & 81.0 & 82.9 & 69.8 & 63.6 & 72   & 71.1 & 60.8 & 72.2   \\
               & LDGNNTrack~\cite{yang2021learning} & \textbf{82.0} & \textbf{83.1} & \textbf{73.4} & 63.5 & \textbf{72.3} & \textbf{71.3} & \textbf{63.5} & \textbf{73.4}   \\
               & Ours                                                                                   & 79.5 & 81.2 & 72.8 & \textbf{66.0} & 69.5 & 70.8 & 61.8 & 72.2   \\ 
\midrule
\multirow{11}{*}{PoseTrack 2018} & STAF~\cite{raaj2019efficient}       & -    & -    & -    & -    & -    & -    & -    & 60.9   \\
               & TML++~\cite{hwang2019pose}      & \textbf{76.0} & 76.9 & 66.1 & 56.4 & 65.1 & 61.6 & 52.4 & 65.7   \\
               & PT\_CPN++~\cite{yu2018multi}  & 68.8 & 73.5 & 65.6 & 61.2 & 54.9 & 64.6 & 56.7 & 64.0   \\
               & LightTrack~\cite{ning2020lighttrack}  & -    & -    & -    & -    & -    & -    & -    & 64.9   \\
               & KeyTrack~\cite{snower202015}   & -    & -    &-    & -    & -    & -    & -    & 66.6   \\
               & PGPT~\cite{bao2020pose}       & 75.4 & 77.3 & 69.4 & 71.5 & 65.8 & \textbf{67.2} & 59.0 & 68.4   \\
               & CombDet~\cite{wang2020combining}    & 74.2 & 76.4 & 71.2 & 64.1 & 64.5 & 65.8 & \textbf{61.9} & 68.7   \\
               & TKMRNet~\cite{zhou2020temporal}    & -    &-    & -    & -    & -    & -    & -    & 68.9   \\
               & SKCTrack~\cite{rafi2020self}  & -    & -    & -    & -    & -    & -    & -    & 69.1   \\
               & LDGNNTrack~\cite{yang2021learning} & 74.3 & 77.6 & 71.4 & 64.3 & 65.6 & 66.7 & 61.7 & \textbf{69.2}   \\
               & Ours                                                                                   & 72.8 & \textbf{77.7} & \textbf{72.4} & \textbf{64.8} & \textbf{66.3} & \textbf{67.2} & 61.1 & \textbf{69.2}   \\
\bottomrule
\end{tabular}}
\end{table*}
\par
\begin{table}[tb]
  \centering
  \caption{Comparison with state-of-the-art multi-person pose estimation and tracking methods on PoseTrack 2017 Test set in terms of AP~(\%) and MOTA~(\%). Results are from PoseTrack 2017 Test Leaderboard.}
  \label{tab:posetrack_pt_ts}
  \setlength{\tabcolsep}{1.9mm}{
  \begin{tabular}{l|c|c|c|c}
   \toprule
   Method      & Wrists AP & Ankles AP &\textbf{Total AP} &  \textbf{Total MOTA} \\
   \midrule
   JointFlow~\cite{xiao2018simple}  &$53.1$ &$50.4$ &$63.4$ &$53.1$ \\
   TML++~\cite{hwang2019pose}      &$60.9$ &$56.0$ &$67.8$ &$54.5$ \\
   FlowTrack~\cite{xiao2018simple}  &$71.5$ &$65.7$ &$74.6$ &$57.8$ \\
   HRNet~\cite{sun2019deep}      &$72.0$ &$67.0$ &\textbf{75.0} &$57.9$ \\
   POINet~\cite{ruan2019poinet}     &$69.5$ &$67.2$ &$72.5$ &$58.4$ \\
   KeyTrack~\cite{snower202015}   &$71.9$ &$65.0$ &$74.0$ &$61.2$ \\
   CombDet~\cite{wang2020combining}    &$69.8$ &$65.9$ &$74.1$ &\textbf{64.1} \\
   Ours        &\textbf{72.7} &\textbf{68.5} &$74.9$ &$63.6$ \\
   \bottomrule
  \end{tabular}}
\end{table}
\textit{\textbf{Analysis of Pose Tracking Pipeline.}} 
With two detectors, HTC~\cite{chen2019hybrid} and YOLOv5~\cite{glenn_jocher_2020_4154370}, built on top of our proposed pose estimation network KCN, we first evaluate the different components of our pose tracking pipeline and quantify how much each of them contributes to the final performance of our proposed method on the validation set of PoseTrack 2018, as shown in Table~\ref{tab:abs_ptp}. 
As we can see, the ID-retrieve module does not enhance performance in multi-person pose estimation, while the Bbox-revision module effectively boosts the performance in both multi-person pose estimation and pose tracking tasks. 
We also observe that the ID-retrieve module improves comparable performance in MOTA with both detectors. 
Meanwhile, the improvement brought by Bbox-revision with HTC detector is significantly surpassing that with YOLOv5 detector in both mAP and MOTA. 
The reason may be that YOLOv5 detector has more missing detections than HTC detector.
\par
Besides, we also evaluate the effectiveness of each component in PTP with our baseline pose estimation network. 
As shown in Table~\ref{tab:abs_ptp_effective}, similarly, both ID-retrieve and Bbox-revision modules boost the pose tracking performance on PoseTrack 2017 and PoseTrack 2018 validation sets. This amply demonstrates the effectiveness of our proposed ID-retrieve module and Bbox-revision.

\textit{\textbf{Effectiveness of KCN and PTP.}} 
In this part, we evaluate the contribution of two components, the Keypoints Confidence Network (KCN) and Pose Tracking Pipeline (PTP), to the final performance of our method on both PoseTrack 2017 and 2018 datasets, as shown in Table~\ref{tab:abs_kcn_ptp}. 
It can be observed that the results improve substantially on both two datasets with our proposed KCN and PTP. 
In particular, KCN delivers significantly higher performance improvement compared to PTP.
The model incorporating KCN and PTP obtain 12.0\%$\sim$16.1\% performance gains over our baseline, which proves the effectiveness of our design.
\par
\subsection{Comparison with State-of-the-art Methods}
We compare our proposed method with the state-of-the-art methods in the multi-person pose estimation and tracking task on PoseTrack 2017 and PoseTrack 2018 datasets. 
\par
\textit{\textbf{Multi-person Pose Estimation.} }
Table~\ref{tab:posetrack_kpf_vs} and Table~\ref{tab:posetrack_pe_vset} and Table~\ref{tab:posetrack_pt_vs} show the comparisons between our proposed method and existing methods on multi-person pose estimation task on the validation sets of PoseTrack 2017 and PoseTrack 2018 datasets. 
Table~\ref{tab:posetrack_kpf_vs} shows the results of multi-person pose estimation with filtering the low confidence keypoints for pose tracking. 
It can be observed that our approach outperforms the most competitive top-down approach, TKMRNet, by 0.5~mAP and outperforms the best bottom-up approach, ST-Embed, by 3.0~mAP on PoseTrack 2017 validation set.
On PoseTrack 2018, our method achieves comparable results with the state-of-the-art method TKMRNet. 
Table~\ref{tab:posetrack_pe_vset} shows the results of multi-person pose estimation in videos where we evaluate the poses without filtering the low confidence keypoints. 
Our method can well recover the human instances that are missed by the detector in videos for multi-person pose estimation. 
It can be seen that our approach achieves the second best performance on PoseTrack 2017 and the best performance on PoseTrack 2018. 
Our method outperforms the most competitive approach, LDGNNTrack, by 0.4~mAP on PoseTrack 2018 validation set. 
\par
Overall, excellence in performance on both PoseTrack 2017 and PoseTrack 2018 validation sets in two tasks, pose estimation with keypoint filtering and multi-person pose estimation without keypoint fully prove the effectiveness of our proposed method.
\par
\textit{\textbf{Mulit-person Pose Tracking.}} 
We compare our method with state-of-the-art multi-person pose tracking methods on both PoseTrack validation sets and the test set. Since the PoseTrack 2018 test set is not available yet on the benchmark server, for the test split we only provide the comparison on PoseTrack 2017 dataset.
As shown in Table~\ref{tab:posetrack_pt_vs} and Table~\ref{tab:posetrack_pt_ts}, our approach outperforms other methods and achieves the best performances on PoseTrack 2018 validation set. On PoseTrack 2017, our approach also achieves an excellent pose tracking performance on both validation set and test set.



%% file: sections/conclusion.tex
In this work, we present a confidence-based novel top-down approach for multi-person pose estimation and tracking task. 
Specifically, we propose a keypoint confidence network and a pose tracking pipeline. 
Compared to the previous methods, the proposed keypoint confidence network considers the availability probability when estimating keypoint confidence, while others only use the location probability. 
We also design a pose tracking pipeline with a Bbox-revision module and an ID-retrieve module to improve the tracking performance. 
The experimental results show that our approach achieves state-of-the-art performance on both PoseTrack 2017 and 2018 datasets.

%% file: bare_jrnl_new_sample4.bbl
\begin{thebibliography}{10}
\providecommand{\url}[1]{#1}
\csname url@samestyle\endcsname
\providecommand{\newblock}{\relax}
\providecommand{\bibinfo}[2]{#2}
\providecommand{\BIBentrySTDinterwordspacing}{\spaceskip=0pt\relax}
\providecommand{\BIBentryALTinterwordstretchfactor}{4}
\providecommand{\BIBentryALTinterwordspacing}{\spaceskip=\fontdimen2\font plus
\BIBentryALTinterwordstretchfactor\fontdimen3\font minus \fontdimen4\font\relax}
\providecommand{\BIBforeignlanguage}[2]{{%
\expandafter\ifx\csname l@#1\endcsname\relax
\typeout{** WARNING: IEEEtran.bst: No hyphenation pattern has been}%
\typeout{** loaded for the language `#1'. Using the pattern for}%
\typeout{** the default language instead.}%
\else
\language=\csname l@#1\endcsname
\fi
#2}}
\providecommand{\BIBdecl}{\relax}
\BIBdecl

\bibitem{liu2018recognizing}
M.~Liu and J.~Yuan, ``Recognizing human actions as the evolution of pose estimation maps,'' in \emph{Proceedings of the IEEE Conference on Computer Vision and Pattern Recognition}, 2018, pp. 1159--1168.

\bibitem{bao2020pose}
Q.~Bao, W.~Liu, Y.~Cheng, B.~Zhou, and T.~Mei, ``Pose-guided tracking-by-detection: Robust multi-person pose tracking,'' \emph{IEEE Transactions on Multimedia}, vol.~23, pp. 161--175, 2020.

\bibitem{8897575}
Y.~Wu, D.~Kong, S.~Wang, J.~Li, and B.~Yin, ``An unsupervised real-time framework of human pose tracking from range image sequences,'' \emph{IEEE Transactions on Multimedia}, vol.~22, no.~8, pp. 2177--2190, 2020.

\bibitem{8089370}
Z.~Liu, Z.~Lin, X.~Wei, and S.-C. Chan, ``A new model-based method for multi-view human body tracking and its application to view transfer in image-based rendering,'' \emph{IEEE Transactions on Multimedia}, vol.~20, no.~6, pp. 1321--1334, 2018.

\bibitem{insafutdinov2017arttrack}
E.~Insafutdinov, M.~Andriluka, L.~Pishchulin, S.~Tang, E.~Levinkov, B.~Andres, and B.~Schiele, ``Arttrack: Articulated multi-person tracking in the wild,'' in \emph{Proceedings of the IEEE Conference on Computer Vision and Pattern Recognition}, 2017, pp. 6457--6465.

\bibitem{wang2020combining}
M.~Wang, J.~Tighe, and D.~Modolo, ``Combining detection and tracking for human pose estimation in videos,'' in \emph{Proceedings of the IEEE Conference on Computer Vision and Pattern Recognition}, 2020, pp. 11\,088--11\,096.

\bibitem{zhou2020temporal}
C.~Zhou, Z.~Ren, and G.~Hua, ``Temporal keypoint matching and refinement network for pose estimation and tracking,'' in \emph{Proceedings of European Conference on Computer Vision}, ser. Lecture Notes in Computer Science, vol. 12367.\hskip 1em plus 0.5em minus 0.4em\relax Springer, 2020, pp. 680--695.

\bibitem{yu2018multi}
D.~Yu, K.~Su, J.~Sun, and C.~Wang, ``Multi-person pose estimation for pose tracking with enhanced cascaded pyramid network,'' in \emph{Proceedings of European Conference on Computer Vision Workshops}, ser. Lecture Notes in Computer Science, vol. 11130, 2018, pp. 221--226.

\bibitem{snower202015}
M.~Snower, A.~Kadav, F.~Lai, and H.~P. Graf, ``15 keypoints is all you need,'' in \emph{Proceedings of the IEEE Conference on Computer Vision and Pattern Recognition}, 2020, pp. 6738--6748.

\bibitem{10.1007/978-3-030-58452-8_13}
N.~Carion, F.~Massa, G.~Synnaeve, N.~Usunier, A.~Kirillov, and S.~Zagoruyko, ``End-to-end object detection with transformers,'' in \emph{Proceedings of the Europe Conference on Computer Vision}, A.~Vedaldi, H.~Bischof, T.~Brox, and J.-M. Frahm, Eds., vol. 12346.\hskip 1em plus 0.5em minus 0.4em\relax Springer, 2020, pp. 213--229.

\bibitem{9932281}
T.~Liang, X.~Chu, Y.~Liu, Y.~Wang, Z.~Tang, W.~Chu, J.~Chen, and H.~Ling, ``Cbnet: A composite backbone network architecture for object detection,'' \emph{IEEE Transactions on Image Processing}, vol.~31, pp. 6893--6906, 2022.

\bibitem{9600874}
T.-I. Chen, Y.-C. Liu, H.-T. Su, Y.-C. Chang, Y.-H. Lin, J.-F. Yeh, W.-C. Chen, and W.~Hsu, ``Dual-awareness attention for few-shot object detection,'' \emph{IEEE Transactions on Multimedia}, pp. 1--1, 2021, doi:~10.1109/TMM.2021.3125195.

\bibitem{9424971}
C.~Zhang, Z.~Li, J.~Liu, P.~Peng, Q.~Ye, S.~Lu, T.~Huang, and Y.~Tian, ``Self-guided adaptation: Progressive representation alignment for domain adaptive object detection,'' \emph{IEEE Transactions on Multimedia}, vol.~24, pp. 2246--2258, 2022.

\bibitem{8445665}
G.~Cheng, J.~Han, P.~Zhou, and D.~Xu, ``Learning rotation-invariant and fisher discriminative convolutional neural networks for object detection,'' \emph{IEEE Transactions on Image Processing}, vol.~28, no.~1, pp. 265--278, 2019.

\bibitem{zou2019object}
Z.~Zou, Z.~Shi, Y.~Guo, and J.~Ye, ``Object detection in 20 years: A survey,'' 2019, arXiv preprint arXiv:1905.05055.

\bibitem{8627998}
Z.-Q. Zhao, P.~Zheng, S.-T. Xu, and X.~Wu, ``Object detection with deep learning: A review,'' \emph{IEEE Transactions on Neural Networks and Learning Systems}, vol.~30, no.~11, pp. 3212--3232, 2019.

\bibitem{9123553}
T.~Kong, F.~Sun, H.~Liu, Y.~Jiang, L.~Li, and J.~Shi, ``Foveabox: Beyound anchor-based object detection,'' \emph{IEEE Transactions on Image Processing}, vol.~29, pp. 7389--7398, 2020.

\bibitem{redmon2016you}
J.~Redmon, S.~Divvala, R.~Girshick, and A.~Farhadi, ``You only look once: Unified, real-time object detection,'' in \emph{Proceedings of the IEEE conference on computer vision and pattern recognition}, 2016, pp. 779--788.

\bibitem{tan2020efficientdet}
M.~Tan, R.~Pang, and Q.~V. Le, ``Efficientdet: Scalable and efficient object detection,'' in \emph{Proceedings of the IEEE/CVF conference on computer vision and pattern recognition}, 2020, pp. 10\,781--10\,790.

\bibitem{girdhar2018detect}
R.~Girdhar, G.~Gkioxari, L.~Torresani, M.~Paluri, and D.~Tran, ``Detect-and-track: Efficient pose estimation in videos,'' in \emph{Proceedings of the IEEE Conference on Computer Vision and Pattern Recognition}, 2018, pp. 350--359.

\bibitem{yang2021learning}
Y.~Yang, Z.~Ren, H.~Li, C.~Zhou, X.~Wang, and G.~Hua, ``Learning dynamics via graph neural networks for human pose estimation and tracking,'' in \emph{Proceedings of the IEEE conference on computer vision and pattern recognition}, 2021, pp. 8074--8084.

\bibitem{pfister2015flowing}
T.~Pfister, J.~Charles, and A.~Zisserman, ``Flowing convnets for human pose estimation in videos,'' in \emph{Proceedings of the IEEE international conference on computer vision}, 2015, pp. 1913--1921.

\bibitem{toshev2014deeppose}
A.~Toshev and C.~Szegedy, ``Deeppose: Human pose estimation via deep neural networks,'' in \emph{Proceedings of the IEEE conference on computer vision and pattern recognition}, 2014, pp. 1653--1660.

\bibitem{newell2016stacked}
A.~Newell, K.~Yang, and J.~Deng, ``Stacked hourglass networks for human pose estimation,'' in \emph{Proceedings of Europe Conference on Computer Vision}, ser. Lecture Notes in Computer Science, vol. 9912.\hskip 1em plus 0.5em minus 0.4em\relax Springer, 2016, pp. 483--499.

\bibitem{wei2016convolutional}
S.-E. Wei, V.~Ramakrishna, T.~Kanade, and Y.~Sheikh, ``Convolutional pose machines,'' in \emph{Proceedings of the IEEE conference on Computer Vision and Pattern Recognition}, 2016, pp. 4724--4732.

\bibitem{cao2017realtime}
Z.~Cao, T.~Simon, S.-E. Wei, and Y.~Sheikh, ``Realtime multi-person 2d pose estimation using part affinity fields,'' in \emph{Proceedings of the IEEE conference on computer vision and pattern recognition}, 2017, pp. 7291--7299.

\bibitem{fang2017rmpe}
H.-S. Fang, S.~Xie, Y.-W. Tai, and C.~Lu, ``Rmpe: Regional multi-person pose estimation,'' in \emph{Proceedings of the IEEE international conference on computer vision}, 2017, pp. 2334--2343.

\bibitem{song2017thin}
J.~Song, L.~Wang, L.~Van~Gool, and O.~Hilliges, ``Thin-slicing network: A deep structured model for pose estimation in videos,'' in \emph{Proceedings of the IEEE conference on computer vision and pattern recognition}, 2017, pp. 4220--4229.

\bibitem{papandreou2017towards}
G.~Papandreou, T.~Zhu, N.~Kanazawa, A.~Toshev, J.~Tompson, C.~Bregler, and K.~Murphy, ``Towards accurate multi-person pose estimation in the wild,'' in \emph{Proceedings of the IEEE conference on computer vision and pattern recognition}, 2017, pp. 4903--4911.

\bibitem{pishchulin2016deepcut}
L.~Pishchulin, E.~Insafutdinov, S.~Tang, B.~Andres, M.~Andriluka, P.~V. Gehler, and B.~Schiele, ``Deepcut: Joint subset partition and labeling for multi person pose estimation,'' in \emph{Proceedings of the IEEE conference on computer vision and pattern recognition}, 2016, pp. 4929--4937.

\bibitem{sun2019deep}
K.~Sun, B.~Xiao, D.~Liu, and J.~Wang, ``Deep high-resolution representation learning for human pose estimation,'' in \emph{Proceedings of the IEEE Conference on Computer Vision and Pattern Recognition}, 2019, pp. 5693--5703.

\bibitem{DBLP:conf/cvpr/TompsonGJLB15}
J.~Tompson, R.~Goroshin, A.~Jain, Y.~LeCun, and C.~Bregler, ``Efficient object localization using convolutional networks,'' in \emph{Proceedings of the IEEE Conference on Computer Vision and Pattern Recognition}.\hskip 1em plus 0.5em minus 0.4em\relax {IEEE} Computer Society, 2015, pp. 648--656.

\bibitem{DBLP:conf/nips/TompsonJLB14}
J.~Tompson, A.~Jain, Y.~LeCun, and C.~Bregler, ``Joint training of a convolutional network and a graphical model for human pose estimation,'' in \emph{Advances in Neural Information Processing Systems 27}, 2014, pp. 1799--1807.

\bibitem{DBLP:conf/eccv/BulatT16}
A.~Bulat and G.~Tzimiropoulos, ``Human pose estimation via convolutional part heatmap regression,'' in \emph{Proceedings of European Conference on Computer Vision}, ser. Lecture Notes in Computer Science, vol. 9911.\hskip 1em plus 0.5em minus 0.4em\relax Springer, 2016, pp. 717--732.

\bibitem{DBLP:conf/cvpr/CarreiraAFM16}
J.~Carreira, P.~Agrawal, K.~Fragkiadaki, and J.~Malik, ``Human pose estimation with iterative error feedback,'' in \emph{Proceedings of the IEEE Conference on Computer Vision and Pattern Recognition,}, 2016, pp. 4733--4742.

\bibitem{DBLP:conf/cvpr/HuR16}
P.~Hu and D.~Ramanan, ``Bottom-up and top-down reasoning with hierarchical rectified gaussians,'' in \emph{Proceedings of IEEE Conference on Computer Vision and Pattern Recognition}, 2016, pp. 5600--5609.

\bibitem{DBLP:conf/cvpr/HuangRSZKFFWSG017}
J.~Huang, V.~Rathod, C.~Sun, M.~Zhu, A.~Korattikara, A.~Fathi, I.~Fischer, Z.~Wojna, Y.~Song, S.~Guadarrama, and K.~Murphy, ``Speed/accuracy trade-offs for modern convolutional object detectors,'' in \emph{Proceedings of the IEEE Conference on Computer Vision and Pattern Recognition}, 2017, pp. 3296--3297.

\bibitem{DBLP:conf/bmvc/RafiLGK16}
U.~Rafi, B.~Leibe, J.~Gall, and I.~Kostrikov, ``An efficient convolutional network for human pose estimation,'' in \emph{Proceedings of the British Machine Vision Conference}, 2016.

\bibitem{xu2022vitpose}
Y.~Xu, J.~Zhang, Q.~Zhang, and D.~Tao, ``Vitpose: Simple vision transformer baselines for human pose estimation,'' \emph{arXiv preprint arXiv:2204.12484}, 2022.

\bibitem{DBLP:conf/fgr/BulatKTP20}
A.~Bulat, J.~Kossaifi, G.~Tzimiropoulos, and M.~Pantic, ``Toward fast and accurate human pose estimation via soft-gated skip connections,'' in \emph{Proceedings of the IEEE International Conference on Automatic Face and Gesture Recognition}, 2020, pp. 8--15.

\bibitem{9677941}
H.~Liu, W.~Liu, Z.~Chi, Y.~Wang, Y.~Yu, J.~Chen, and T.~Jin, ``Fast human pose estimation in compressed videos,'' \emph{IEEE Transactions on Multimedia}, pp. 1--1, 2022, doi:~10.1109/TMM.2022.3141888.

\bibitem{9858008}
G.~Kim, H.~Kim, K.~Kong, J.-W. Song, and S.-J. Kang, ``Human body-aware feature extractor using attachable feature corrector for human pose estimation,'' \emph{IEEE Transactions on Multimedia}, pp. 1--11, 2022.

\bibitem{8064661}
G.~Ning, Z.~Zhang, and Z.~He, ``Knowledge-guided deep fractal neural networks for human pose estimation,'' \emph{IEEE Transactions on Multimedia}, vol.~20, no.~5, pp. 1246--1259, 2018.

\bibitem{DBLP:conf/cvpr/PishchulinJATS12}
L.~Pishchulin, A.~Jain, M.~Andriluka, T.~Thorm{\"{a}}hlen, and B.~Schiele, ``Articulated people detection and pose estimation: Reshaping the future,'' in \emph{Proceedings of the IEEE Conference on Computer Vision and Pattern Recognition}, 2012, pp. 3178--3185.

\bibitem{he2017mask}
K.~He, G.~Gkioxari, P.~Doll{\'a}r, and R.~Girshick, ``Mask r-cnn,'' in \emph{Proceedings of the IEEE international conference on computer vision}, 2017, pp. 2961--2969.

\bibitem{DBLP:conf/cvpr/ChenWPZYS18}
Y.~Chen, Z.~Wang, Y.~Peng, Z.~Zhang, G.~Yu, and J.~Sun, ``Cascaded pyramid network for multi-person pose estimation,'' in \emph{proceedings of the IEEE Conference on Computer Vision and Pattern Recognition}, 2018, pp. 7103--7112.

\bibitem{xiao2018simple}
B.~Xiao, H.~Wu, and Y.~Wei, ``Simple baselines for human pose estimation and tracking,'' in \emph{Proceedings of the Europe Conference on Computer Vision}, 2018, pp. 466--481.

\bibitem{DBLP:conf/cvpr/GkioxariHGM14}
G.~Gkioxari, B.~Hariharan, R.~B. Girshick, and J.~Malik, ``Using k-poselets for detecting people and localizing their keypoints,'' in \emph{Proceedings of the IEE Conference on Computer Vision and Pattern Recognition,}, 2014, pp. 3582--3589.

\bibitem{DBLP:conf/cvpr/ChenY15}
X.~Chen and A.~L. Yuille, ``Parsing occluded people by flexible compositions,'' in \emph{Proceedings of the IEEE Conference on Computer Vision and Pattern Recognition}, 2015, pp. 3945--3954.

\bibitem{DBLP:conf/eccv/InsafutdinovPAA16}
E.~Insafutdinov, L.~Pishchulin, B.~Andres, M.~Andriluka, and B.~Schiele, ``Deepercut: {A} deeper, stronger, and faster multi-person pose estimation model,'' in \emph{Proceedings of the European Conference on Computer Vision}, ser. Lecture Notes in Computer Science, B.~Leibe, J.~Matas, N.~Sebe, and M.~Welling, Eds., vol. 9910.\hskip 1em plus 0.5em minus 0.4em\relax Springer, 2016, pp. 34--50.

\bibitem{DBLP:conf/eccv/IqbalG16}
U.~Iqbal and J.~Gall, ``Multi-person pose estimation with local joint-to-person associations,'' in \emph{Proceedings of the Europe Conference on Computer Vision Workshops}, ser. Lecture Notes in Computer Science, vol. 9914, 2016, pp. 627--642.

\bibitem{DBLP:conf/cvpr/KreissBA19}
S.~Kreiss, L.~Bertoni, and A.~Alahi, ``Pifpaf: Composite fields for human pose estimation,'' in \emph{Proceedings of the IEEE Conference on Computer Vision and Pattern Recognition}, 2019, pp. 11\,977--11\,986.

\bibitem{DBLP:journals/tits/KreissBA22}
K.~Sven, B.~Lorenzo, and A.~Alexandre, ``Openpifpaf: Composite fields for semantic keypoint detection and spatio-temporal association,'' \emph{IEEE Transactions on Intelligent Transportation Systems}, vol.~23, no.~8, pp. 13\,498--13\,511, 2022.

\bibitem{raaj2019efficient}
Y.~Raaj, H.~Idrees, G.~Hidalgo, and Y.~Sheikh, ``Efficient online multi-person 2d pose tracking with recurrent spatio-temporal affinity fields,'' in \emph{Proceedings of the IEEE Conference on Computer Vision and Pattern Recognition}, 2019, pp. 4620--4628.

\bibitem{jin2019multi}
S.~Jin, W.~Liu, W.~Ouyang, and C.~Qian, ``Multi-person articulated tracking with spatial and temporal embeddings,'' in \emph{Proceedings of the IEEE Conference on Computer Vision and Pattern Recognition}, 2019, pp. 5664--5673.

\bibitem{DBLP:conf/eccv/GuoTLCLW18}
H.~Guo, T.~Tang, G.~Luo, R.~Chen, Y.~Lu, and L.~Wen, ``Multi-domain pose network for multi-person pose estimation and tracking,'' in \emph{Proceedings of the Europe Conference on Computer Vision Workshops}, ser. Lecture Notes in Computer Science, vol. 11130.\hskip 1em plus 0.5em minus 0.4em\relax Springer, 2018, pp. 209--216.

\bibitem{he2016deep}
K.~He, X.~Zhang, S.~Ren, and J.~Sun, ``Deep residual learning for image recognition,'' in \emph{Proceedings of the IEEE Conference on Computer Vision and Pattern Recognition}, 2016, pp. 770--778.

\bibitem{lin2017focal}
T.-Y. Lin, P.~Goyal, R.~Girshick, K.~He, and P.~Doll{\'a}r, ``Focal loss for dense object detection,'' in \emph{Proceedings of the International Conference on Computer Vision}, 2017, pp. 2980--2988.

\bibitem{oks}
``{O}bject {K}eypoint {S}imilarity,'' \url{https://cocodataset.org/#keypoints-eval}, accessed: 2023-05-12.

\bibitem{teed2020raft}
Z.~Teed and J.~Deng, ``Raft: Recurrent all-pairs field transforms for optical flow,'' in \emph{Proceedings of the Europe Conference on Computer Vision}, ser. Lecture Notes in Computer Science, vol. 12347.\hskip 1em plus 0.5em minus 0.4em\relax Springer, 2020, pp. 402--419.

\bibitem{chen2019hybrid}
K.~Chen, J.~Pang, J.~Wang, Y.~Xiong, X.~Li, S.~Sun, W.~Feng, Z.~Liu, J.~Shi, W.~Ouyang \emph{et~al.}, ``Hybrid task cascade for instance segmentation,'' in \emph{Proceedings of the IEEE Conference on Computer Vision and Pattern Recognition}, 2019, pp. 4974--4983.

\bibitem{mmdetection}
K.~Chen, J.~Wang, J.~Pang, Y.~Cao, Y.~Xiong, X.~Li, S.~Sun, W.~Feng, Z.~Liu, J.~Xu, Z.~Zhang, D.~Cheng, C.~Zhu, T.~Cheng, Q.~Zhao, B.~Li, X.~Lu, R.~Zhu, Y.~Wu, J.~Dai, J.~Wang, J.~Shi, W.~Ouyang, C.~C. Loy, and D.~Lin, ``{MMDetection}: Open mmlab detection toolbox and benchmark,'' \emph{arXiv preprint arXiv:1906.07155}, 2019.

\bibitem{glenn_jocher_2020_4154370}
\BIBentryALTinterwordspacing
G.~Jocher, A.~Stoken, J.~Borovec, NanoCode012, ChristopherSTAN, L.~Changyu, Laughing, tkianai, A.~Hogan, lorenzomammana, yxNONG, AlexWang1900, L.~Diaconu, Marc, wanghaoyang0106, ml5ah, Doug, F.~Ingham, Frederik, Guilhen, Hatovix, J.~Poznanski, J.~Fang, L.~Yu, changyu98, M.~Wang, N.~Gupta, O.~Akhtar, PetrDvoracek, and P.~Rai, ``{ultralytics/yolov5: v3.1 - Bug Fixes and Performance Improvements},'' Oct. 2020. [Online]. Available: \url{https://doi.org/10.5281/zenodo.4154370}
\BIBentrySTDinterwordspacing

\bibitem{jin2017towards}
S.~Jin, X.~Ma, Z.~Han, Y.~Wu, W.~Yang, W.~Liu, C.~Qian, and W.~Ouyang, ``Towards multi-person pose tracking: Bottom-up and top-down methods,'' in \emph{Proceedings of the IEEE International Conference on Computer Vision Workshops}, vol.~2, no.~3, 2017, p.~7.

\bibitem{zhu2017multi}
X.~Zhu, Y.~Jiang, and Z.~Luo, ``Multi-person pose estimation for posetrack with enhanced part affinity fields,'' in \emph{Proceedings of the IEEE International Conference on Computer Vision Workshops}, vol.~7, 2017, p. 4321.

\bibitem{xiu2018pose}
Y.~Xiu, J.~Li, H.~Wang, Y.~Fang, and C.~Lu, ``Pose flow: Efficient online pose tracking,'' \emph{arXiv preprint arXiv:1802.00977}, 2018.

\bibitem{hwang2019pose}
J.~Hwang, J.~Lee, S.~Park, and N.~Kwak, ``Pose estimator and tracker using temporal flow maps for limbs,'' in \emph{Proceedings of the IEEE International Joint Conference on Neural Networks}, 2019, pp. 1--8.

\bibitem{ning2020lighttrack}
G.~Ning, J.~Pei, and H.~Huang, ``Lighttrack: A generic framework for online top-down human pose tracking,'' in \emph{Proceedings of the IEEE Conference on Computer Vision and Pattern Recognition workshops}, 2020, pp. 1034--1035.

\bibitem{rafi2020self}
U.~Rafi, A.~Doering, B.~Leibe, and J.~Gall, ``Self-supervised keypoint correspondences for multi-person pose estimation and tracking in videos,'' in \emph{Proceedings of the Europe Conference on Computer Vision}.\hskip 1em plus 0.5em minus 0.4em\relax Springer, 2020, pp. 36--52.

\bibitem{bertasius2019learning}
G.~Bertasius, C.~Feichtenhofer, D.~Tran, J.~Shi, and L.~Torresani, ``Learning temporal pose estimation from sparsely-labeled videos,'' \emph{arXiv preprint arXiv:1906.04016}, 2019.

\bibitem{ruan2019poinet}
W.~Ruan, W.~Liu, Q.~Bao, J.~Chen, Y.~Cheng, and T.~Mei, ``Poinet: pose-guided ovonic insight network for multi-person pose tracking,'' in \emph{Proceedings of the ACM International Conference on Multimedia}, 2019, pp. 284--292.

\end{thebibliography}
